\documentclass[12pt]{iopart}

\usepackage[dvips]{graphicx}
\begin{document}

\title[]
{Stationary probability density of stochastic search processes in global optimization}

\author{Arturo Berrones}

\address{Posgrado en Ingenier\' \i a de Sistemas,
Facultad de Ingenier\' \i a Mec\'anica y El\'ectrica
Universidad Aut\'onoma de Nuevo Le\'on
AP 126, Cd. Universitaria, 
San Nicol\'as de los Garza, NL 66450, M\'exico}
\ead{arturo@yalma.fime.uanl.mx}
\begin{abstract}
A method for the construction of approximate analytical expressions for the
stationary marginal densities of general stochastic search processes is 
proposed. By the marginal densities, regions of the search space that 
with high probability contain the global optima can be readily defined.
The density estimation procedure involves a controlled number of linear
operations, with a computational cost per iteration that grows linearly with 
problem size.
\end{abstract}



\maketitle

\section{Diffusion and Global Optimization}

Stochastic strategies for optimization 
are essential to many of the heuristic techniques used to deal with complex, 
unstructured global optimization problems. 
Methods like simulated annealing \cite{anneal,anneal2,anneal3,anneal4} 
and evolutionary algorithms \cite{ga,eva,pelikan}, have
proven to be valuable tools, capable of give good quality solutions at a 
relatively small computational effort.
In spite of their success, these approaches present a major drawback, namely
the absence of valid bounds on the obtained solutions.
A common feature of deterministic global optimization algorithms is the
progressive reduction of the domain space until the global optimum has been 
found with arbitrary accuracy \cite{floudas,horst}. 
An analogous property for stochastic algorithms has been largely lacking.
In this contribution is introduced a method for the 
estimation of the asymptotic probability density of a general stochastic 
search process in global optimization problems.
The convergence of the estimated density can be clearly assessed, and with the
help of this density, reliable bounds for the location of the global optimum 
are derived. 
The procedure involves linear operations only, and a well defined number of
evaluations of the given cost function. The presented results indicate that 
by the proposed approach, regions of the search space
can be discarded on a probabilistic basis. This property
may be implemented in a variety of ways in order to improve existing or
develop new optimization algorithms, and open the door for the construction
of probabilistic optimality certificates in large scale nonlinear optimization problems.

The roots of stochastic search methods can be traced back to the Metropolis algorithm \cite{metropolis}, introduced in 
the early days of scientific computing to simulate the
evolution of a physical system to thermal equilibrium. This process is the base
of the simulated annealing technique \cite{anneal}, which makes use of the convergence to
a global minimum in configurational energy observed in physical systems 
at thermal equilibrium as the temperature
goes to zero. The method presented in this contribution is rooted in similar
physical principles as those on which simulated annealing and related
algorithms \cite{anneal,fplearning,anneal2,anneal3} are based.
However, in contrast with other approaches, 
the proposed method considers a density of points instead of 
Markov transitions of individual points. Moreover, the main goal of the
proposed approach is not the convergence to global minima as a randomness parameter
is reduced, but the approximation of the probability density after an infinitely long exploration
time of the search space, keeping a fixed randomness.

Consider the minimization of a cost function of the form $V(x_1, x_2, ..., x_n, ..., x_N)$ 
with a search space defined over 
$L_{1,n} \leq x_n \leq L_{2,n}$.  A stochastic  search process for this problem is modeled by

\begin{eqnarray} \label{langevin}
\dot{x}_n = - \frac{\partial V}{\partial x_n} + \varepsilon(t) ,
\end{eqnarray} 

\noindent
where $\varepsilon(t)$ is an additive noise with zero mean. Equation (\ref{langevin}), 
known as Langevin equation in the Statistical Physics literature \cite{risken,vankampen}, 
captures the basic properties of a general stochastic search strategy. 
Under an uncorrelated Gaussian noise with constant strength, Eq. (\ref{langevin})  
represents a search by diffusion, while a noise strength that is slowly varying in time gives a 
simulated annealing. Notice that 
when choosing an external noise of infinite amplitude, 
the dynamical influence of the cost function over the exploration process is lost, leading to a blind search. The model given by Eq.(\ref{langevin}) 
can be also interpreted as an overdamped
nonlinear dynamical system composed by $N$ interacting particles. 
The temporal evolution of the probability density of such a system 
in the presence of an additive Gaussian white noise, is described 
by a linear differential equation, the Fokker -- Planck equation \cite{risken,vankampen},

\begin{eqnarray}\label{fp}
\dot{p} = 
\frac{\partial }{\partial x} \left [ \frac{\partial V}{\partial x} p\right ]
+D\frac{\partial ^{2} p}{\partial x^{2}}
\end{eqnarray}

\noindent
where $D$ is a constant, called diffusion constant, that is proportional to the noise
strength. The direct use of Eq. (\ref{fp})
for optimization or deviate generation purposes would imply 
the calculation of high dimensional integrals.
It results numerically much less demanding to perform the 
following one dimensional projection of Eq. (\ref{fp}).
Under very general conditions (e. g., the absence of infinite 
cost values), the equation (\ref{fp}) has a stationary solution
over a search space with reflecting boundaries \cite{risken, grasman}. 
The stationary conditional probability density
satisfy the one dimensional Fokker -- Planck equation

\begin{eqnarray}\label{sfp}
D\frac{\partial p(x_n | \{ x_{j\neq n} = x_j^{*} \} )}{\partial x_n}
+p(x_n | \{ x_{j\neq n} = x_j^{*} \} ) \frac{\partial V}{\partial x_n} = 0 .
\end{eqnarray}

\noindent
An important consequence of Eq. (\ref{sfp}) is that the marginal $p(x_n)$ can be
sampled by drawing points from the conditional $p(x_n | \{ x_{j\neq n} = x_j^{*} \} )$
via a Gibbs sampling \cite{geman}. It is now shown how, due to the linearity of the
Fokker -- Planck equation, a particular form of Gibbs sampling can be constructed, 
such that its not only possible to sample the marginal density, but to give 
an approximate analytical expression for it.
From Eq. (\ref{sfp}) follows 
a linear second order 
differential equation for the cumulative distribution $y(x_n | \{ x_{j\neq n} = 
x_j^{*} \} ) = \int _{-\infty}^{x_n} 
p(x^{'}_{n} | \{ x_{j\neq n} = x_j^{*} \} ) dx^{'}_{n}$,

\begin{eqnarray}\label{sfpm}
\frac{d^{2}y}{dx_{n}^{2}}+\frac{1}{D}\frac{\partial V}{\partial x_n}\frac{dy}{dx_n}
= 0 ,\\ \nonumber
\\ \nonumber
y(L_{1,n})=0, \quad y(L_{2,n})=1 .
\end{eqnarray}

\noindent
Random deviates can be drawn from the density $p(x_n | \{ x_{j\neq n} = x_j^{*} \} )$ by the fact 
that $y$ is an uniformly distributed random variable
in the interval $y \in [0, 1]$. Viewed as a function of the random variable $x_n$, 
$y(x_n | \{ x_{j\neq n} \})$ can be approximated through a 
linear combination of functions from a complete set 
that satisfy the boundary conditions in the interval of interest,

\begin{eqnarray}\label{set}
\hat{y}(x_n | \{ x_{j\neq n} \})=\sum_{l=1}^{L} a_l \varphi _l  ( x_n  ) .
\end{eqnarray}

\noindent
Choosing for instance, a basis in which $\varphi _ l ( 0 ) = 0$, the $L$ coefficients
are uniquely defined by the evaluation of Eq. (\ref{sfpm}) in $L-1$ interior points. In this way, the 
approximation of $y$ is performed by solving a set of $L$ linear algebraic equations, involving
$L-1$ evaluations of the derivative of $V$.

The proposed procedure is based on the iteration of the following steps:

\noindent
{\bf 1)} Fix the variables $x_{j\neq n} = x_j^{*}$ and approximate $y(x_n | \{ x_{j\neq n} \})$
 by the use of formulas
(\ref{sfpm}) and (\ref{set}).

\noindent
{\bf 2)} By the use of $\hat{y}(x_n | \{ x_{j\neq n} \})$ 
construct a lookup table in order to
generate a deviate 
$x_n^{*}$ drawn from the stationary distribution
$p(x_n | \{ x_{j\neq n} = x_j^{*} \})$.

\noindent
{\bf 3)} Update $x_n = x_n^{*}$ and repeat the procedure for a new variable $x_{j \neq n}$.

\noindent
The iteration of the three steps above give an algorithm
for the estimation of the equilibrium distribution of the
stochastic search process described by Eq. (\ref{langevin}).
A convergent representation for $p(x_n)$ is obtained after taking the average
of the coefficients $a$'s in the expansion (\ref{set})
over the iterations. 
In order to see this, consider the
expressions for the marginal density and the conditional distribution,

\begin{eqnarray}
p(x_n) = 
\int p(x_n | \{ x_{j\neq n} \}) 
p(\{ x_{j\neq n} \}) d\{ x_{j\neq n} \} ,
\end{eqnarray}

\begin{eqnarray}
y(x_n | \{ x_{j\neq n} \}) =
\int_{-\infty}^{x_n} p(x^{'}_{n} | \{ x_{j\neq n} \} ) dx^{'}_{n} .
\end{eqnarray}

\noindent
From the last two equations follow that the marginal $y(x_n)$ is given by the 
expected value of the conditional $y(x_n | \{ x_{j\neq n} \} )$ 
over the set $\{x_{j\neq n}\}$,

\begin{eqnarray}
y(x_n) = E_{\{ x_{j\neq n} \}} [y(x_n | \{ x_{j\neq n} \} )] .
\end{eqnarray}

\noindent
All the information on the set $\{x_{j\neq n}\}$ is stored in the coefficients
of the expansion (\ref{set}). Therefore 

\begin{eqnarray}\label{setav}
\left< \hat{y} \right> =\sum_{l=1}^{L} \left< a_l \right> \varphi _l  ( x_n  ) 
\to y(x_n) ,
\end{eqnarray}

\noindent where the brackets represent the average over the iterations of the density
estimation procedure.

Previous preliminary applications of the density estimation method on
the generation of suitable populations of initial points 
for optimization algorithms can be 
found in \cite{berrones}. In the next section the capabilities of the proposed algorithm for the
construction of reliable probabilistic bounds is tested on several benchmark 
unconstrained examples and in a family of well known constrained NP-hard problems.  

\section{Examples}

The fundamental parameters of the density estimation procedure, $L$ and $D$, 
have a clear meaning, which is very helpful for their selection. The diffusion 
constant ``smooth'' the density. This is evident by taking the limit $D \to \infty$ in
Eq. (\ref{sfpm}), which imply an uniform density in the domain. The number of base functions
$L$, on the other hand, defines the algorithms capability to ``learn'' more or less
complicated density structures. Therefore, for a given $D$, 
the number $L$ should be at least large enough to assure 
that the estimation algorithm will generate valid distributions $y(x_n | \{ x_{j\neq n} \})$. 
A valid distribution
should be a monotone increasing continuos function that satisfy the boundary conditions.
The parameter $L$ ultimately determines the computational cost of the procedure, because 
at each iteration a system of size $\propto L$ of linear algebraic equations must be solved
$N$ times.
Therefore, the user is able to control the computational cost through the interplay of the
two basic parameters: for a larger $D$ a smoother density should be estimated, 
so a lesser $L$ can be used.

The density estimation algorithm is tested on the following benchmark unconstrained problems:

Schwefel:
\begin{eqnarray}\nonumber
N = 6,\quad f = 418.9829N - \sum_{n=1}^{N} x_n sin(\sqrt{|x_n|}), \\ \nonumber
-500 \leq x_n \leq 500, \quad solution: \quad x^{*} = (420.9687, ...,420.9687), f(x^{*}) = 0. 
\end{eqnarray}

Levy No. 5:
\begin{eqnarray}\nonumber
N = 2 \quad f = \sum_{i=1}^{5} i cos ((i - 1)x_{1} + i) \sum_{j=1}^{5} j cos ((j + 1)x_{2} + j) \\ \nonumber
+(x_{1} + 1.42513)^{2} + (x_{2} + 0.80032)^{2}, \\ \nonumber
-10 \leq x_n \leq 10, \quad solution: \quad x^{*} = (-1.3068, -1.4248), f(x^{*}) = -176.1375.
\end{eqnarray}

Booth: 
\begin{eqnarray}\nonumber
N = 2, \quad  f = (x_1 + 2x_2 - 7)^{2} + (2x_1 + x_2 - 5)^{2}, \\ \nonumber
-10 \leq x_n \leq 10, \quad solution: \quad x^{*} = (1, 3), f(x^{*}) = 0.
\end{eqnarray}

Colville: 
\begin{eqnarray}\nonumber
N = 4, \quad  f = 100 (x_2 - x_1)^2 + (1-x_1)^2 + 90(x_4 - x_3)^2 + (1-x_3)^2 \\ \nonumber
+ 10.1 ((x_2 -1)^2 + (x_4 -1)^2) + 19.8(x_2 -1)(x_4 -1), \\ \nonumber
-10 \leq x_n 10, \quad solution: \quad x^{*} = (1, ..., 1), f(x^{*}) = 0. 
\end{eqnarray}

Rosenbrock: 
\begin{eqnarray}\nonumber
N = 20, \quad  f = \sum_{n=1}^{N} 100 (x_{n+1}-x_{n}^{2})^{2} + (x_n-1)^{2}, \\ \nonumber
-10 \leq x_n \leq 10, \quad solution: \quad x^{*} = (1, ..., 1), f(x^{*}) = 0.
\end{eqnarray}

For the experiments, the following specific form of the expansion (\ref{set})
has been used,

\begin{eqnarray}\label{set2}
\hat{y}=\sum_{l=1}^{L} a_l sin \left ( 
(2l-1)\frac{\pi (x_n - L_{1,n} )}{2(L_{2,n}-L_{1,n})}
\right ),
\end{eqnarray}

\noindent
so the size of the algebraic system of equations is $L-1$. 
The linear system has been solved by a LU decomposition routine \cite{nr}.
The gradients have been calculated
numerically, with two cost function evaluations per derivative. In this way, the total
number of cost function evaluations per iteration goes like $2(L-1)N$.

\begin{figure*}[h]
\begin{center}
{\includegraphics[width=.4\textwidth]{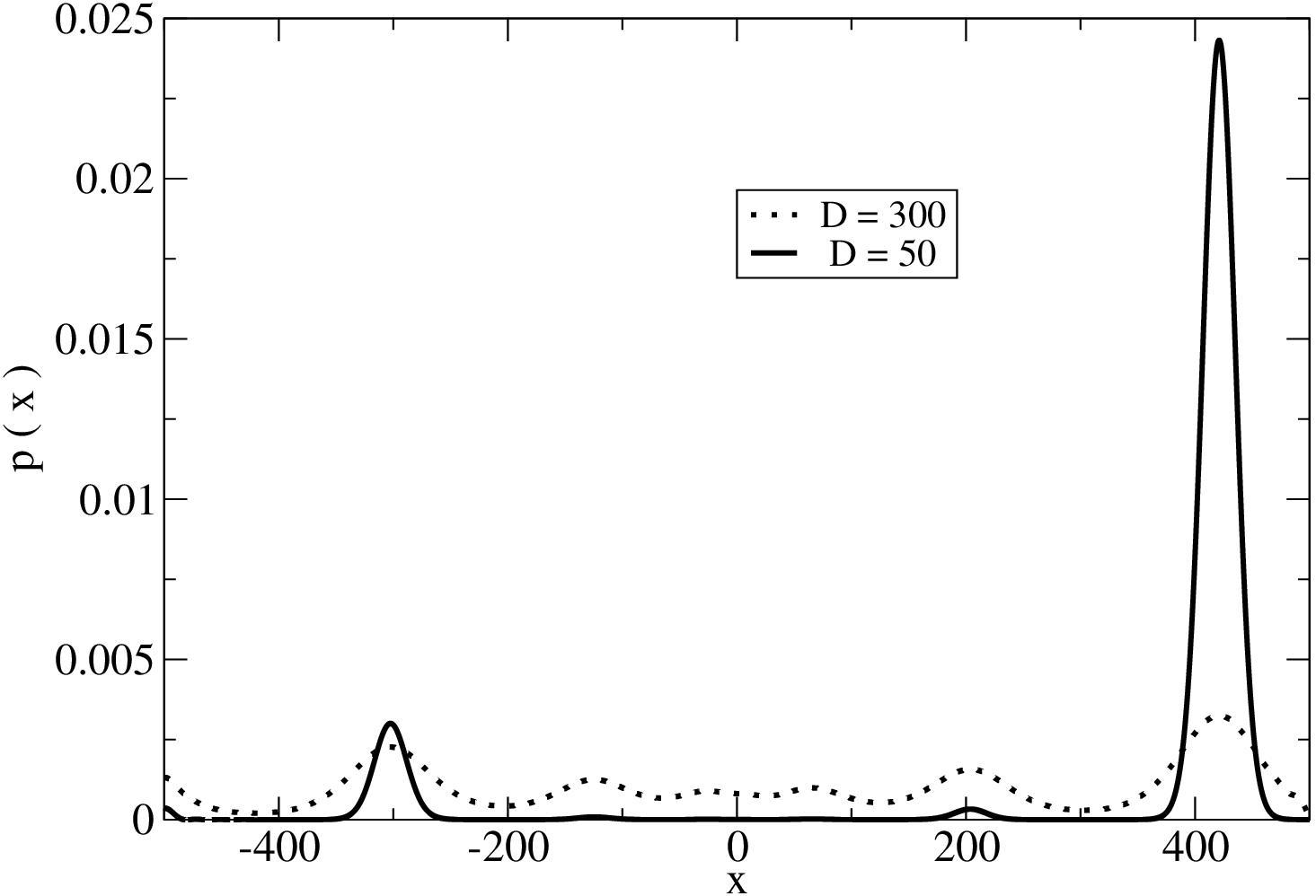}}
\caption{Density estimation for the variable $x_1$ of the 
Schwefel function. Two different diffusion constants have been considered, taking $L = 100$
in both cases. 
In the two cases the density clearly represents the structure of the cost function. 
The density is sharply peaked around the optimal value for the lesser $D$.}\label{schwefel}
\end{center}
\end{figure*}

In Fig. \ref{schwefel} two different pairs $L, D$ have been considered in the study of the
Schwefel problem. This problem has a second best minimum at a relatively large distance
of the global optimum. This is reflected on the estimated densities, but at small $D$ a clear
distiction between the two regions is made. The Schwefel problem is 
an example of a separable function, that is, a function given by a linear
combination of terms, where each term involves a single variable.
Separable problems generate
an uncoupled dynamics of the stochastic search described by Eq. (\ref{langevin}). Because of
this fact, the estimation algorithm converges in only one iteration for separable
problems. 
The Schwefel example also illustrates that the density estimation algorithm works well on 
functions that are not derivable in some points. This is a consequence of the
finite number of gradient evaluations required by the procedure.

\vskip 0.5cm

\begin{figure*}[h]
\begin{center}
{\includegraphics[width=.4\textwidth]{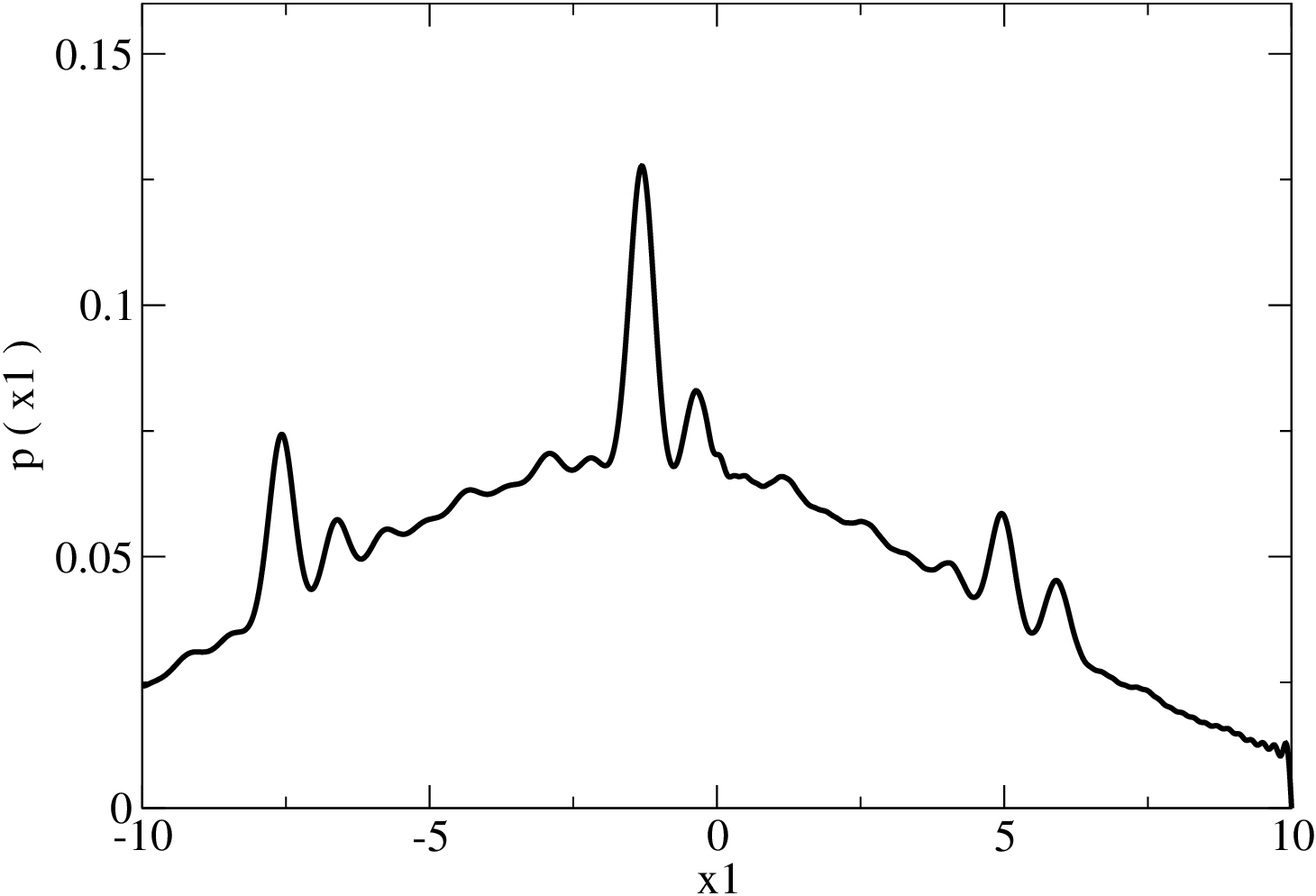}}
\hskip 0.1cm
{\includegraphics[width=.4\textwidth]{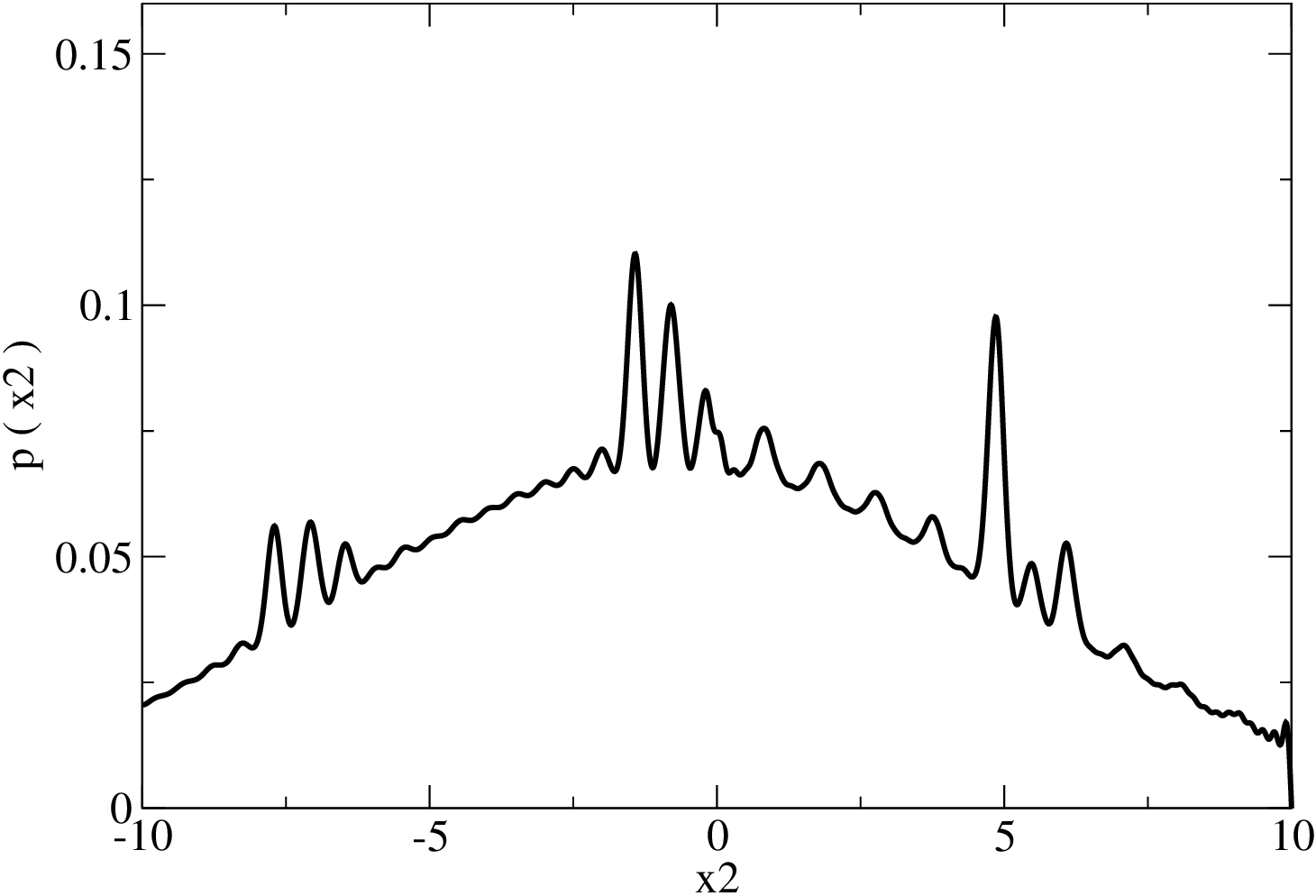}}
\caption{Probability densities associated to the Levy No. 5 problem, 
using the parameter values $L = 200$, $D = 70$ and $M = 300$.
The densities maxima are at coordinates $(-1.3, -1.42)$.}\label{levy}
\end{center}
\end{figure*}

In contrast to the Schwefel function, the stochastic search process associated to
the Levy No. 5 problem represents a coupled nonlinear dynamics. 
Despite that this problem has about $760$ local minima \cite{pso},
the estimation algorithm shows 
good convergence properties, as is illustrated in
Fig. \ref{levy}.

\vskip 0.5cm

The two previous examples show how, once that the estimation algorithm as
attained convergence, is possible to
define a region
of the search space in which with high probability the global optimum is located.
This concept is more sistematically studied 
through the introduction of normalized distances.
The distance normalized with respect to the search space $L_{1,n} \leq x_n \leq L_{2,n}$ 
between two points $x$ and $x^{*}$ is defined by

\begin{eqnarray}
distance = \sqrt{ \frac {(x_1 - x^{*}_1)^{2}+ ... + (x_N - x^{*}_{N})^{2}} 
{(L_{1,1} - L_{2,1})^{2}+ ... + (L_{1,N} - L_{2,N})^{2}}}.
\end{eqnarray}

\begin{figure*}[h]
\begin{center}
{\includegraphics[width=.4\textwidth]{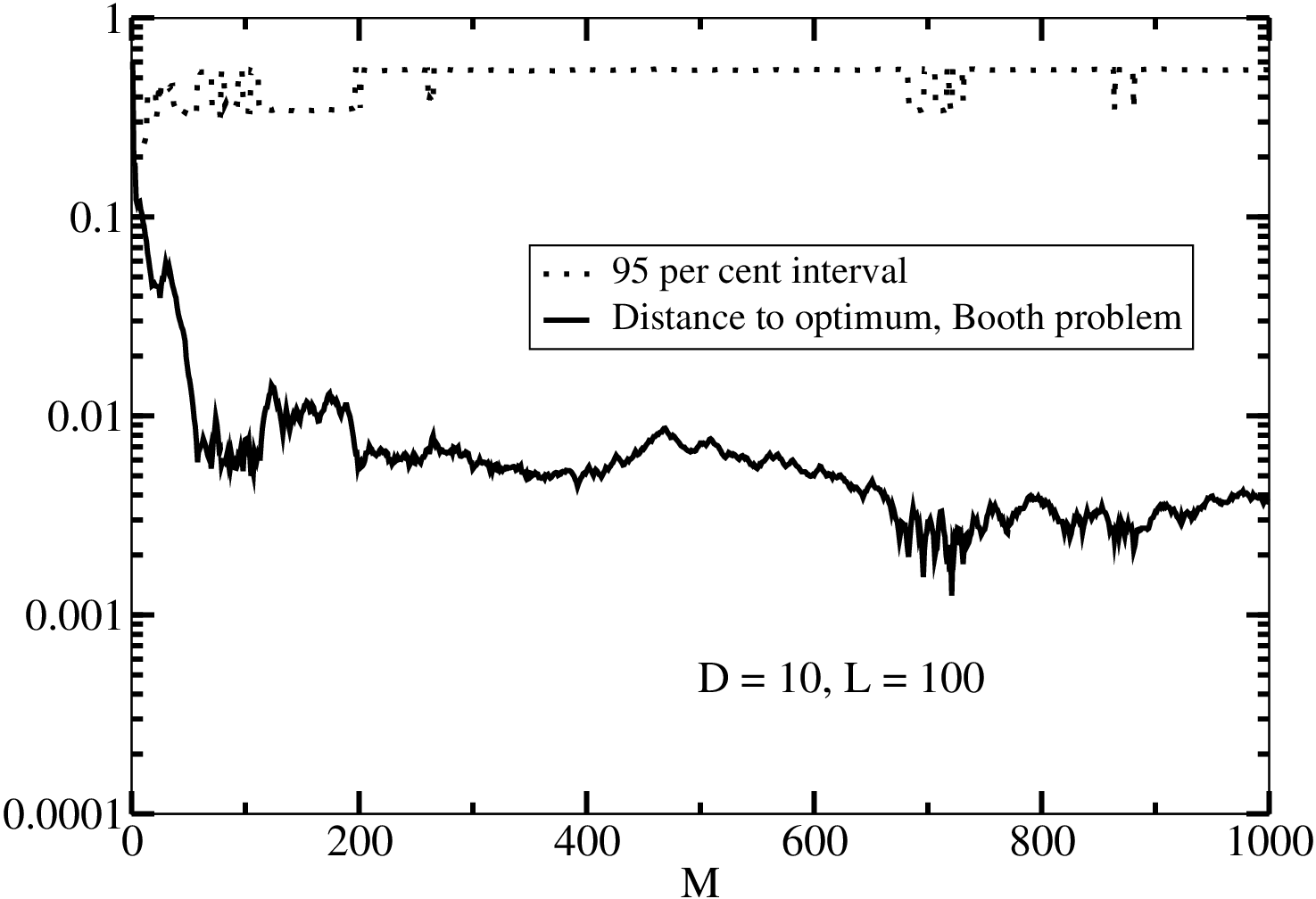}}
\hskip 0.1cm
{\includegraphics[width=.4\textwidth]{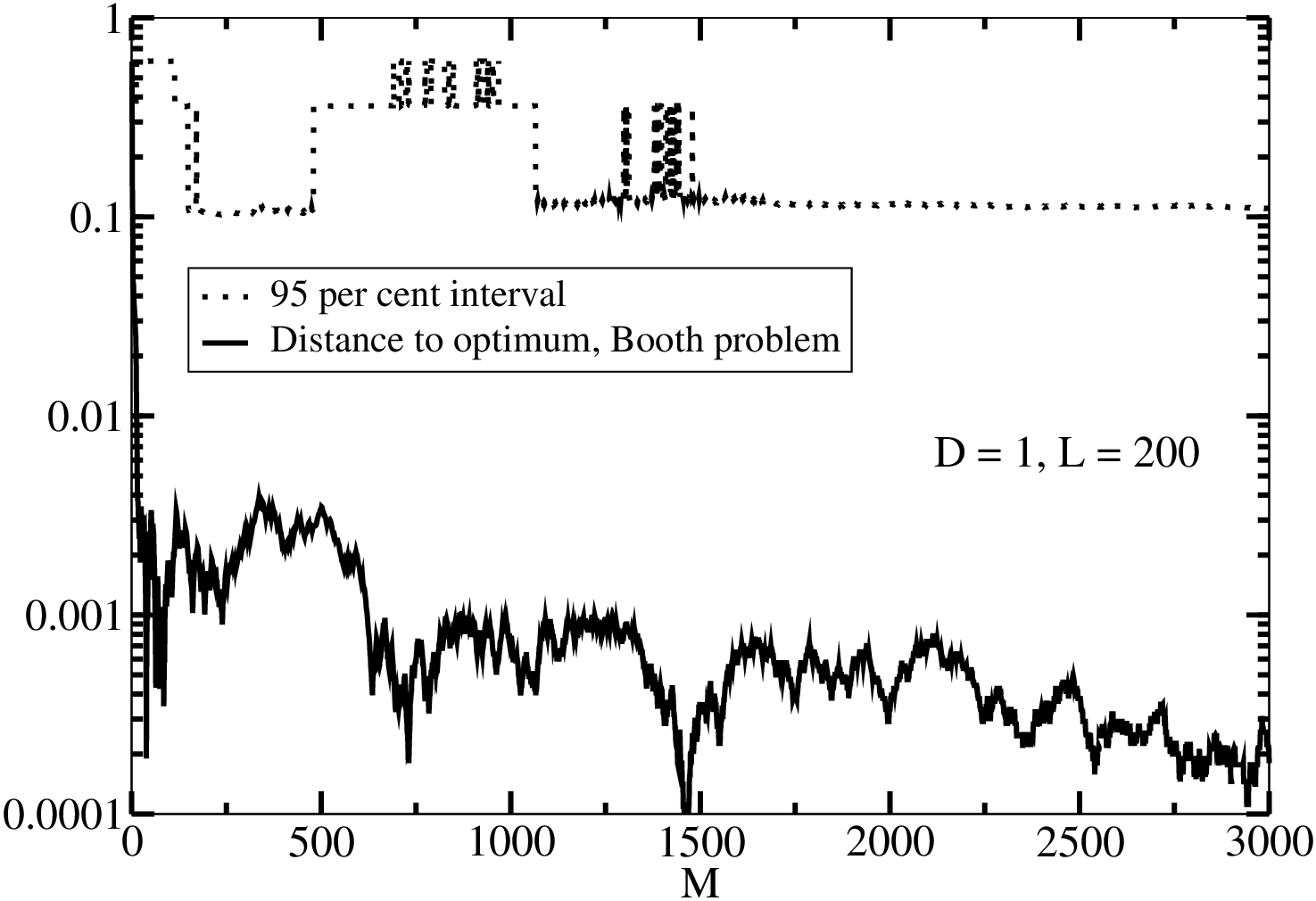}}
\caption{Density estimation for the Booth problem. Semi -- log scale has been used.}\label{booth}
\end{center}
\end{figure*}

\begin{figure*}[h]
\begin{center}
{\includegraphics[width=.4\textwidth]{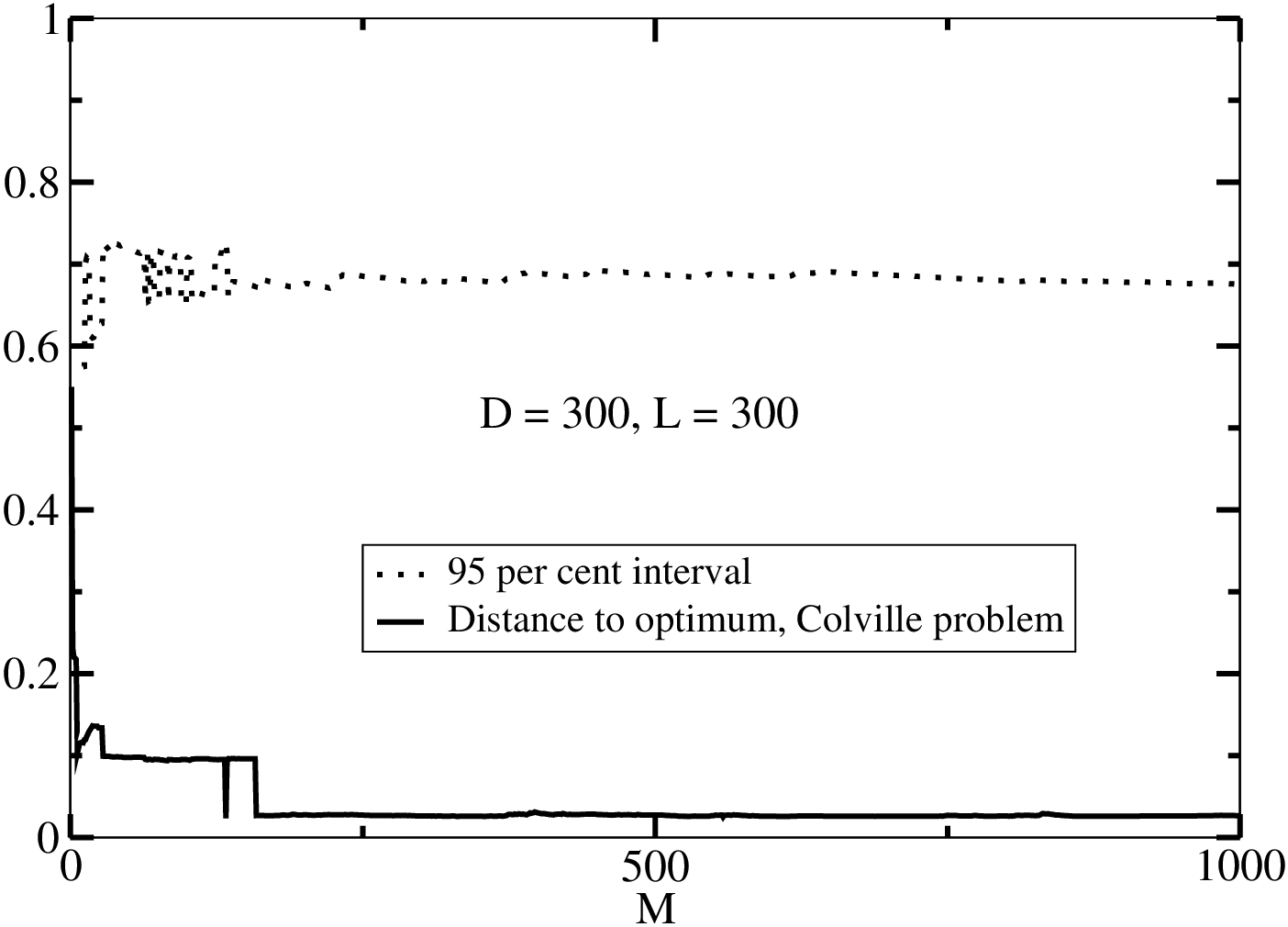}}
\caption{Density estimation for the Colville problem.}\label{colville}
\end{center}
\end{figure*}

\begin{figure*}[h]
\begin{center}
{\includegraphics[width=.4\textwidth]{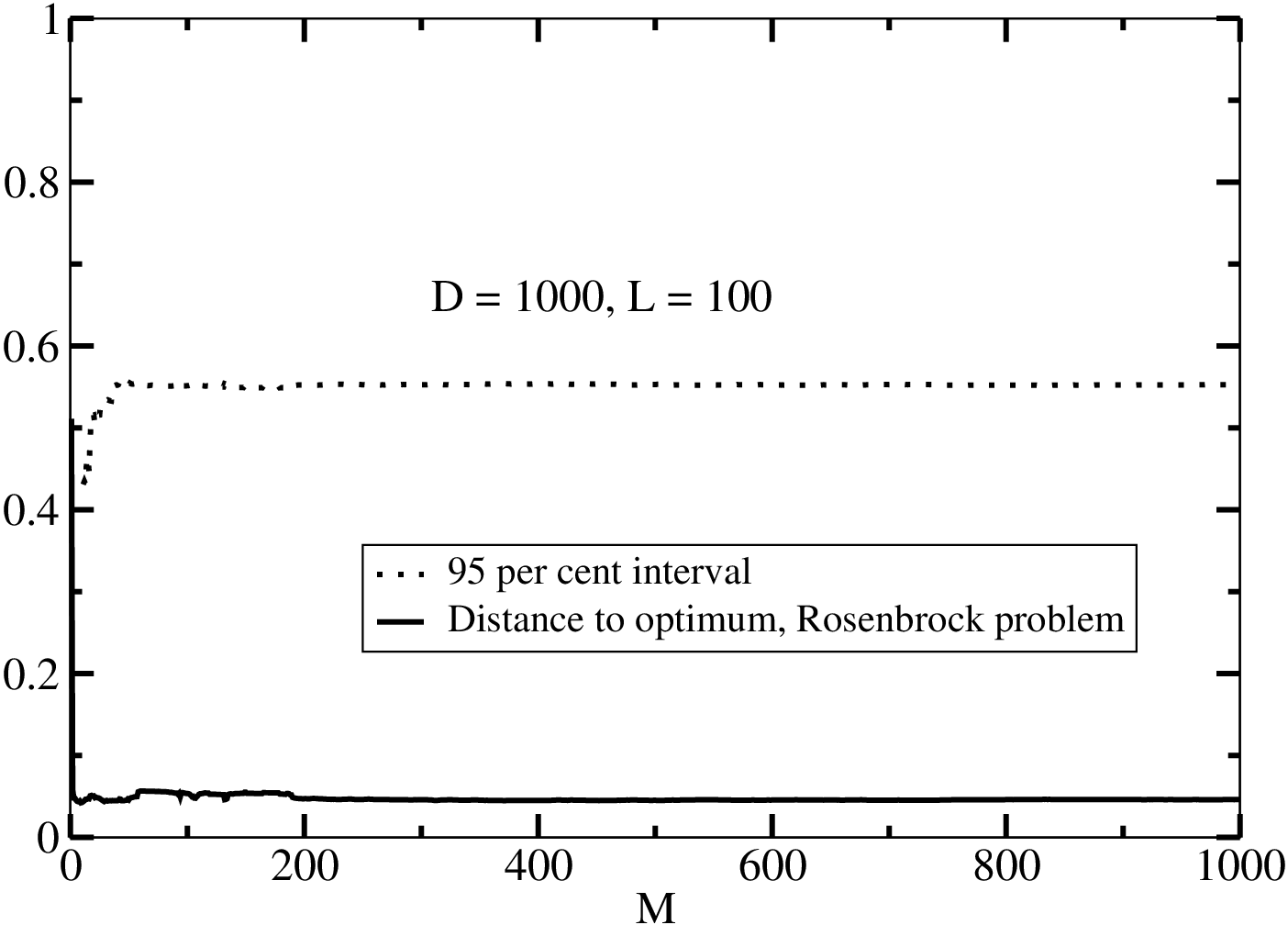}}
\caption{Density estimation for the Rosenbrock problem.}\label{rosen}
\end{center}
\end{figure*}

\noindent
Two measures written in terms of normalized
distances are presented in the examples of Figures \ref{booth}, \ref{colville} and \ref{rosen}:
{\bf i)} The distance between the global optimum and the point in which the density is maximum.
{\bf ii)} The length of the $95 \%$ probability interval around the point of maximum
probability. 

\vskip 0.5cm

\vskip 0.5cm

\vskip 0.5cm

Under general conditions a Gibbs sampling 
displays geometric convergence \cite{roberts}. In the presented experiments the running time has been
chosed such that the $95 \%$ probability interval 
differ in less than $0.01$ between succesive iterations. Additionally, several control runs
from different and independent starting conditions have been performed, indicating convergence 
to the same corresponding region within the predefined accuracy.
As expected for a Gibbs sampling, the density appears to
contract to a region of space that is independent of the staring point \cite{canty}.
The numerical realizations suggest that convergence is 
attained at a few hundreds of iterations for all of the examples, even 
for the $20$ -- dimensional Rosenbrock problem, which as been reported to be difficult to solve
by stochastic heuristics like genetic algorithms \cite{laguna}. 

The numerical experiments show that the 
global optimum is contained in a region close to the
point of maximum probability, and that this region gets more sharp
as $D$ decreases. 
A straighforward application of
this behavior would be, for instance, on simulated 
annealing -- type algorithms. Starting with a large diffusion constant, search regions
can be iteratively discarded. By the use of the density estimation method,
a probability measure is associated with each region. In this way the user can define 
a certain level of precision in the search. 
Several statistical quantities
can be readily calculated like measures of confidence, for instance
probability intervals or characteristic fluctuation sizes.

Because the estimation algorithm depends
on linear operations only, additional nonlinearities in the cost function
can be treated with essentialy the same efficiency, giving more freedom and
flexibility in modeling. For instance, 
the application of the density estimation algorithm 
to constrained problems can be done in a very direct manner through the addition 
of suitable ``energy barriers'' 
(or more precisely, ``force barriers'').
These barriers don't need to be very large. Their main purpose is not to define
prohibited regions, but only regions with low probability.
The original constrained problem is transformed to an unconstrained
cost function with additional nonlinearities.
Of course, the design of adequate barriers may be 
a difficult problem -- dependent task. However, at least for some problems the approach
seems to be straightforward.
This is illustrated on the classical NP-hard
knapsack problem \cite{kp}. 
It is well known that many standard instances
of the knapsack model can be efficiently solved by exact methods \cite{kp},
which makes it an ideal example for
experimentation with the density estimation algorithm.
The knapsack problem is formulated as

\begin{eqnarray}
min \quad - \sum_{n=1}^{N} q_n x_n \\ \nonumber
s. t. \quad \sum_{n=1}^{N} w_n x_n \leq c, \\ \nonumber
-x^{2}_{n} + x_n \leq 0, \quad 0 \leq x_n \leq 1,
\end{eqnarray}

\noindent
where $q_n$, $w_n$ and $c$ are positive numbers. 
The quadratic constraint is equivalent to the usual restriction to binary variables.
The following transformation is proposed,

\begin{eqnarray}
min \quad - \sum_{n=1}^{N} q_n x_n \\ \nonumber 
+ k_0 \sum_{n=1}^{N}
\frac{1}{1+exp( -b_0 [-x^{2}_{n} + x_n] )} \\ \nonumber
+ k_1 \frac{ exp\left( b_1 [\sum_{n=1}^{N} w_n x_n - c] \right) - 1}
{exp\left( -b_2 [\sum_{n=1}^{N} w_n x_n - c] \right) + 1 } \\ \nonumber
s. t. \quad 0 \leq x_n \leq 1,
\end{eqnarray}

\begin{figure*}[h]
\begin{center}
{\includegraphics[width=.3\textwidth]{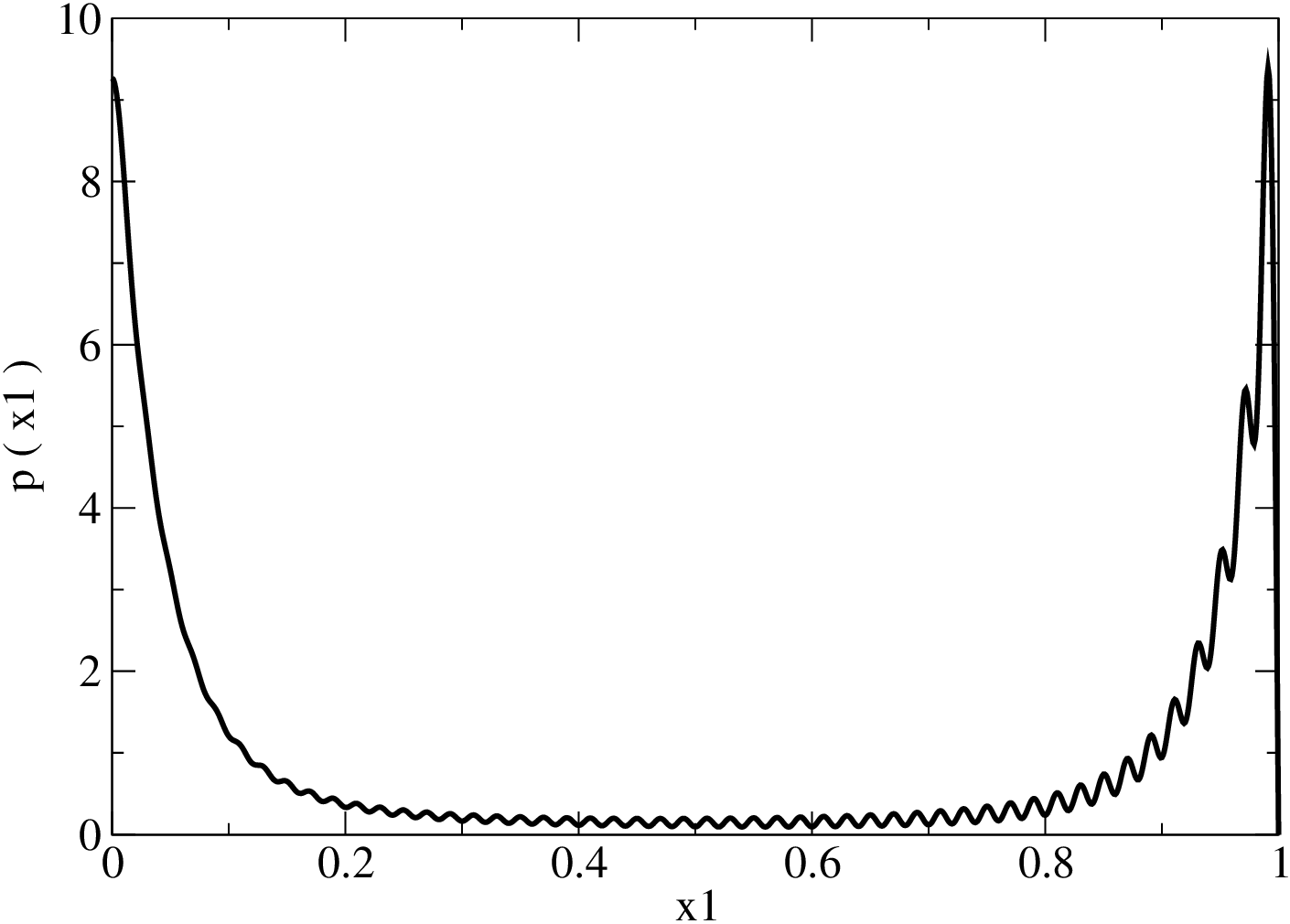}}
\hskip 0.15cm
{\includegraphics[width=.3\textwidth]{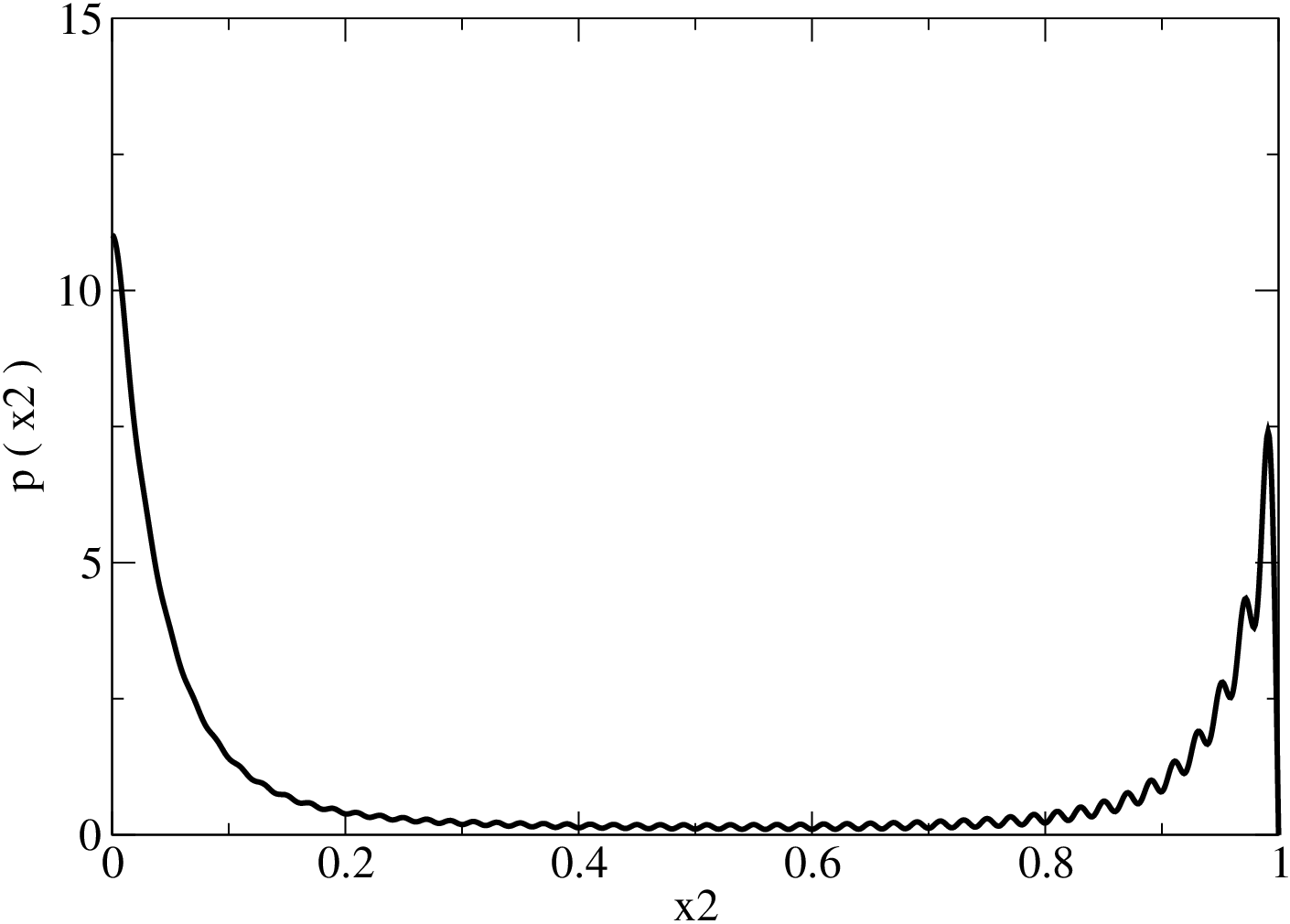}}
\hskip 0.15cm
{\includegraphics[width=.3\textwidth]{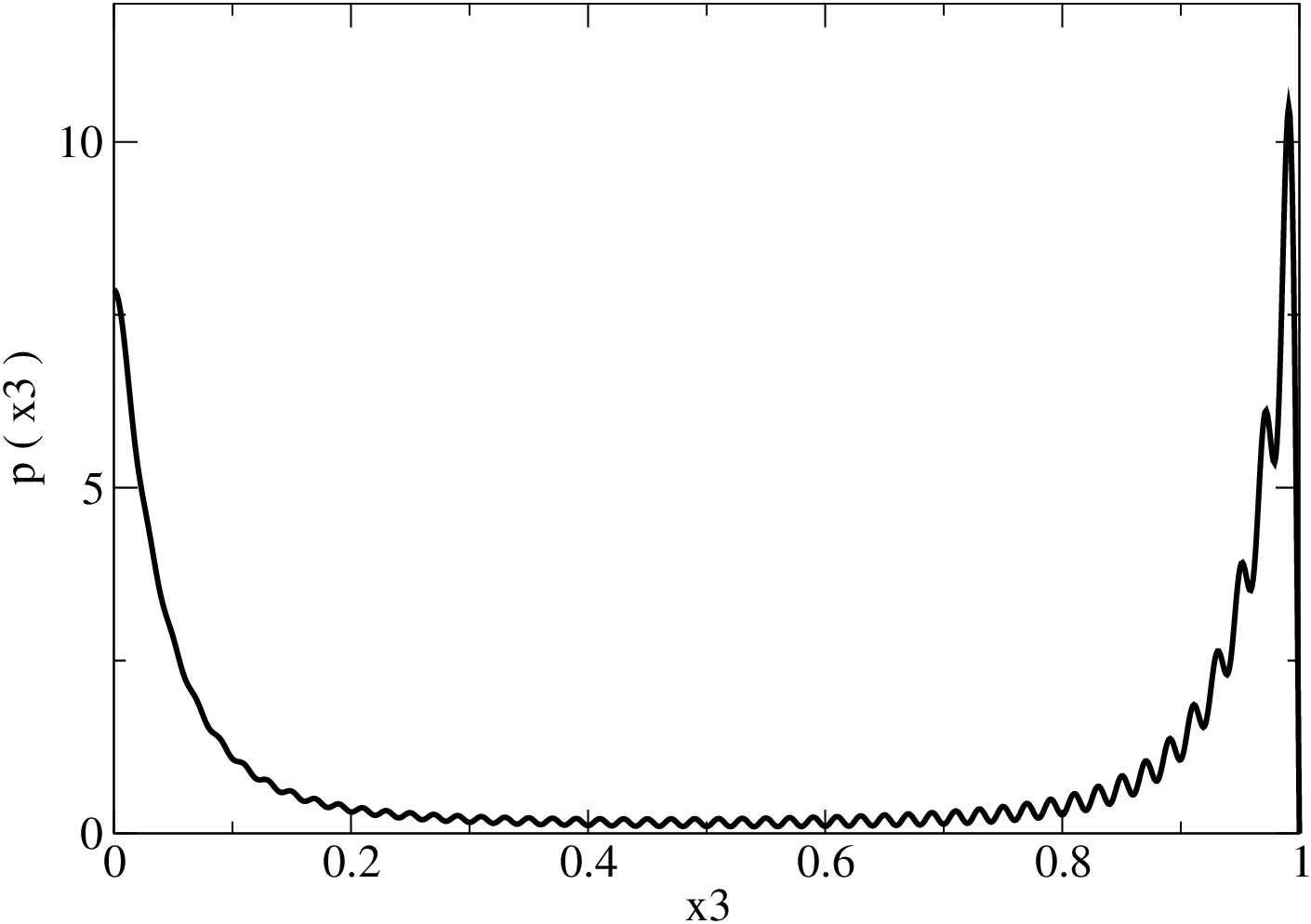}}
\caption{Probability densities associated to a three
dimensional instance of a knapsack problem.}\label{ilustKp}
\end{center}
\end{figure*}

\noindent
For illustrative purposes, consider the instance $q=(2,3,5)$, $w=(3,5,7)$, $c=10$ of the
knapsack problem. By inspection, the solution is given by $x=(1,0,1)$. In Fig. \ref{ilustKp}
typical densities produced by the estimation algorithm for this instance are shown. 
The selection of the parameters has been done after the performance of short runs, measuring
the effects of each of the nonlinear terms. The parameters have been tuned such that:
{\bf a)} The first nonlinear term alone produce symmetric densities peaked in the neighborhood of 
$\{ 0, 1 \}$. {\bf b)} The addition of the second nonlinear term and the original cost function
generate densities in which the configurations with maximum probabilty satisfy the 
$\sum_{n=1}^{N} w_n x_n \leq c$ constraint.
Notice that the configuration that corresponds to the global optimum has the
maximum probability.

\vskip 0.5cm

\begin{figure*}[h]
\begin{center}
{\includegraphics[width=.4\textwidth]{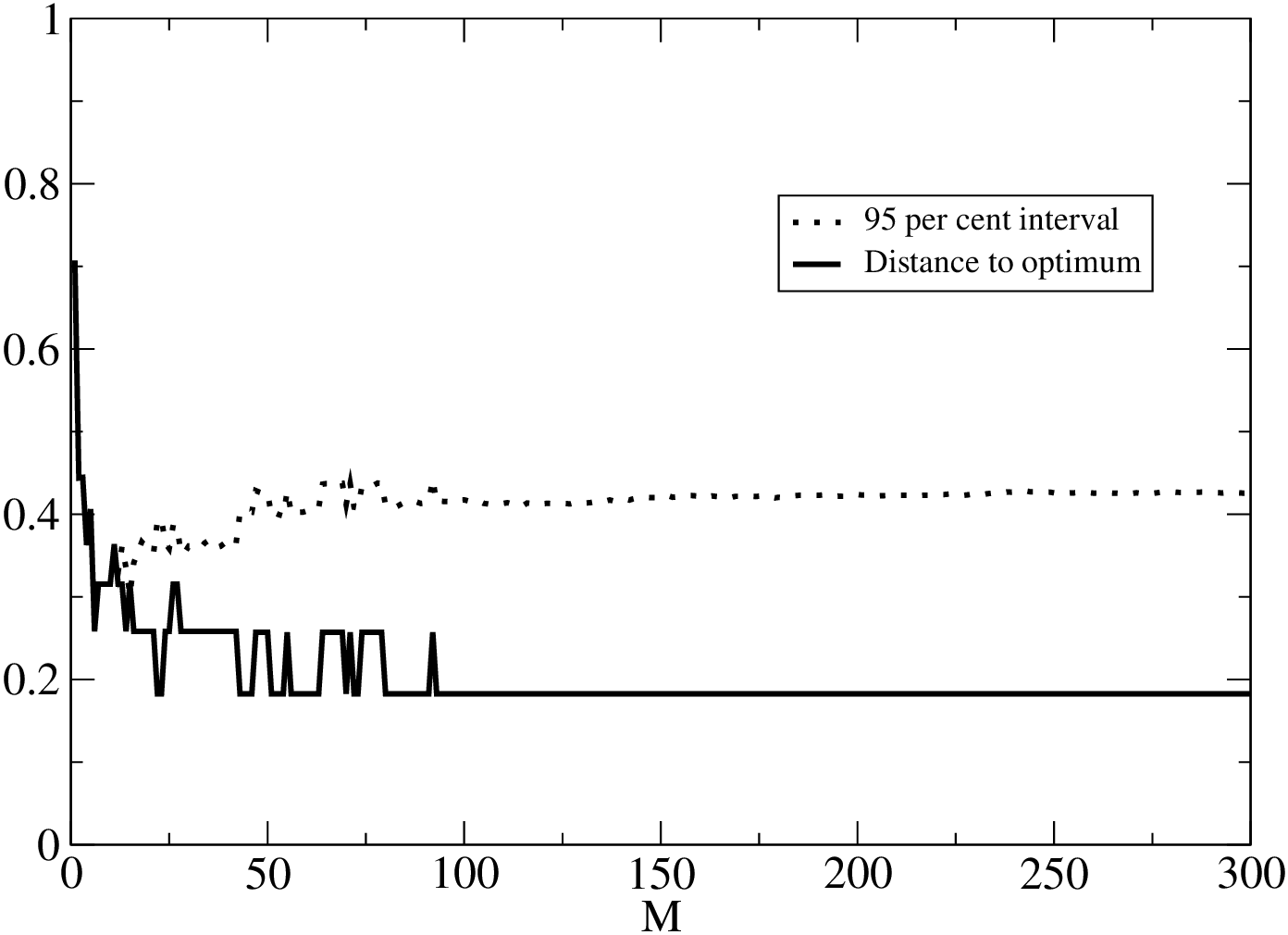}}
\caption{The density estimation of an instance of a knapsack problem with $30$ variables.}\label{Kp}
\end{center}
\end{figure*}

A larger example is presented in Fig. \ref{Kp}. 
The exact solution has been calculated with the branch and bound algorithm
supplied in the GNU Linear Programming Kit \cite{gnu}.
The instance has been generated by taking $w_n$ uniformily distributed in an
interval $[1, R]$, and

\begin{eqnarray} \label{instance}
q_n = w_n + (\varepsilon _n -1)\frac{R}{100} + \frac{R}{10}, \\ \nonumber
\varepsilon _n \quad uniform \quad deviate \quad in \quad [-1, 1] ,
\end{eqnarray}

\noindent
which imply strong linear correlations between $q_n$ and $w_n$. Instances of this
kind are relevant to real management problems in which the return of
an investment is proportional to the sum invested within small variations \cite{kp}.
It has been argued that these type of instances fall in a category which is close
to the ``worst case'' scenario for exact algorithms \cite{kp}. Despite of that, the
estimation method converge to densities in which the optimum is contained
in a region with high probability. Moreover, from the definition of the normalized 
distance, follows that the closest integers to the 
elements of the vector that represent the point of maximum probability differ in $\sim 1$
positions from the exact solution.

\vskip 0.5cm

Three independent realizations of the numerical experiment
over instances generated by Eq. (\ref{instance}) have been performed, varying the orders
of magnitude of $R$ and $c$.  
The results are
summarized in Table 1, indicating the number of $0 \leftrightarrow 1$ flips
that the normalized distances imply.

\begin{table*}[h] 
\caption{Density estimation on instances of the knapsack problem with $30$ variables. The final two 
columns indicate the resulting $95 \%$ intervals and gaps to optimum like normalized distances/flips.}
\begin{tabular*}{\hsize}{@{\extracolsep{\fill}}rrrrrrrrrrrrr}
\hline
\multicolumn1c{\small $R$}&\multicolumn1c{\small $c$}&
\multicolumn1c{\small $k_0$}&\multicolumn1c{\small $k_1$}&
\multicolumn1c{\small $b_0$}&\multicolumn1c{\small $b_1$}&\multicolumn1c{\small $b_2$}& 
\multicolumn1c{\small $M$}&\multicolumn1c{\small $L$}&\multicolumn1c{\small $D$}&
\multicolumn1c{Interval \quad Gap}
\cr
\hline
\small
$10$ & \small $100$ & \small$10$ & \small$7.1$ & \small$10$ & \small$0.01$ & \small$2 b_1$ & \small$300$ & \small$100$ & \small$1$ & \small$0.425$/\small$5$ \quad \small$0.185$/\small$1$ \cr
\small$100$ & \small$500$ & \small$100$ & \small$37.0$ & \small$10$ & \small$0.001$ & \small$3 b_1$ & \small$300$ & \small$100$ & \small$15$ & \small$0.432$/\small$6$ \quad \small$0.363$/\small$4$ \cr
\small$1000$ & \small$3000$ & \small$1000$ & \small$315.0$ & \small$10$ & \small$0.0001$ & \small$3 b_1$ & \small$300$ & \small$100$ & \small$100$ & \small$0.433$/\small$6$ \quad \small$0.257$/\small$2$ \cr 
\hline
\end{tabular*}
\end{table*}

It should be remarked that, although the instances of the
knapsack model discussed in this section
are quickly solvable by exact algorithms, no reference to the
particular structure of the original problem has been used for the density estimation. In fact, 
the problem has been treated like a highly nonlinear cost function of $30$ variables.  

\section{Heuristics Based on Stationary Density Estimation}

The potential benefits of the stationary density estimation algorithm as a tool for the 
construction of new heuristics for
high dimensional
global optimization problems is illustrated through the following greedy random search procedure:

\noindent
1) Run an iteration of the stationary density estimation algorithm.

\noindent
2) An initial best point is given by the global maximum of the estimated density 

\noindent
3) Define a population of $N+1$ points. One point is the current best solution and other is
the current point with maximum probability density. 
The remaining
$N-1$ points are randomly drawn from an uniform distribution centered around the best point. 
For each dimension, the corresponding uniform distribution has a length equal to the typical
fluctuation size given by the estimated density.

\noindent
4) Run a downhill simplex routine. The starting conditions are given by the simplex defined by  
the points generated at step 3 as vertices.

\noindent
5) If from step 4 results a point which improves the best known objective value, then update the best point.

\noindent
6) Run an iteration of the density estimation algorithm.

\noindent
7) Go to step 3.

From an evolutionary perspective, the above procedure acts at two different levels. At a short
time scale finite populations evolve locally in a very greedy fashion. 
On the other hand, 
large changes on the population composition are dictated by a long time scale dynamics, which 
is consistent with the learned information about the global cost landscape. This information is gained
through the approximation of the long term statistical density of a diffusive search process.

The short time scale exploration of the solution space is dominated by the downhill simplex
method, which is a deterministic search based on function evaluations of a simplex vertices \cite{nelder-mead}.  
In a $N$ dimensional search space, a population of $N+1$ points defined by the corresponding
$N+1$ vertices evolve under simple geometric transformations, namely reflection, expansion and contraction.
At each iteration, a new trial point is generated by the image of the worst point in the simplex (reflection).
If the new point is better than all other vertices, the simplex expands in its direction. If the
trial point improves the worst point, a reflection from the new worst point is performed. A contraction
step is made when the worst point is at least as good as the reﬂected point. In this way the simplex
eventually surrounds a local minimum. 
In the experiments presented here, the implementation of the downhill 
simplex given by \cite{nr} has been used, with a fractional decrease of cost value of at least
$10e-4$ as termination criteria. A maximum of $50000$ function evaluations in the downhill
simplex routine is allowed.

From a given reference point $x^{(best)}$, an initial simplex for each call to the downhill simplex routine is defined
through the characteristic length scales $\lambda_n$, as $x^{(i)} = x^{(best)} + \lambda_n e_n$ where
the $e_n$ are $N$ unit vectors. The estimated long term density provides a vertex (the point with maximum likelihhood
at the current stage) and most importantly, typical fluctuation sizes, denoted by $\sigma_n$. 
These are given by the first two moments 
of the estimated density,

\begin{eqnarray} \label{sigma}
\sigma_n = \sqrt{\left < x_{n}^{2} \right > - \left < x_{n} \right >^{2}} 
\end{eqnarray}  

Due to the simple form of the expansion of the estimated density, all the integrals over the variables
domain that are needed for moment calculation are performed analitically. The resulting expressions are
finite sums with $L$ terms. 

\noindent The typical fluctuation sizes (\ref{sigma}) provide a natural definition for the characteristic
length scales $\lambda_n$, in the sense expressed in step (3) of the greedy diffusive search described above.

\begin{figure*}[h]
\begin{center}
{\includegraphics[width=.4\textwidth]{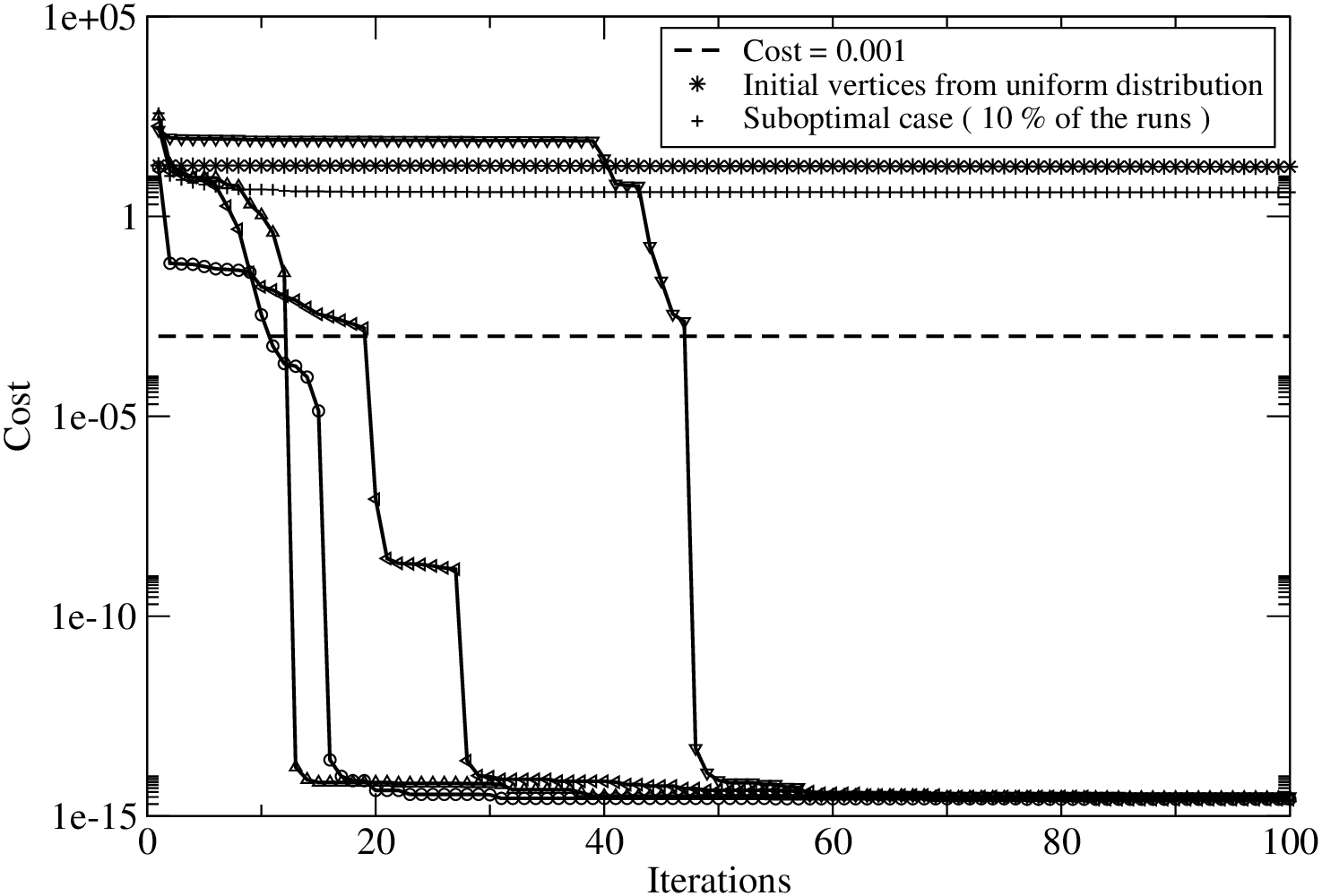}}
\caption{Greedy stochastic search for the Rosenbrock problem.
The parameter values of the density estimation are $L=30$ and $D=10000$.}\label{rosen2}
\end{center}
\end{figure*}

For illustration purposes, consider the Rosenbrock test function.
In Fig.\ref{rosen2} are presented some
plots of the beahavior of our greedy stochastic search for the Rosenbrock problem of $N = 20$ variables. 
The graphs represent the cost function values over 
successive iterations. Four samples from a total of $100$ runs are plotted. The 
result of a version of the algorithm in which an uniform density over the search space is used instead of 
the estimated long term density is also plotted. The success in finding the optimum is defined by a gap size 
with the known global optimum lesser than $0.001$. Each run consist of $100$ iterations.
Over the total number of runs of the greedy stochastic search, $90 \%$ have been successful, and 
the $100 \%$ of the runs outperform the search based on uniform distributions. 
An $80 \%$ of the runs have been successful in less than $48$ iterations and $30 \%$ in less
than $13$ iterations. 
For the $13$ iterations cases, an average number of $28080$ cost function evaluations was needed.
It should be remarked that the $20$-dimensional Rosenbrock test function has been
reported to be extremely difficult to solve by randomized optimization algorithms. 
For instance, in an experiment similar to the one presented here, 
it has been reported in \cite{hu} a success rate of zero after $400000$ function evaluations for 
Simulated Annealing, Cross--Entropy and Model Reference Adaptive Search. 
On the other hand, in \cite{laguna} is reported that
Genetic Algorithms and Scatter Search methods
are unable to succeed in the $20$-dimensional Rosenbrock
problem after $50000$ cost function evaluations.

\vskip 0.5cm

A more exhaustive experimentation with possible heuristics based on the stationary density 
estimation algorithm is in progress.

\section{Conclusion}

The presented results strongly suggest that the proposed density estimation
algorithm can be used to construct probabilistic bounds on the location
of global optima for large classes of problems. The density
estimation is performed in a well defined number of elementary operations. 
The developed theory and the 
numerical experiments indicate that any given desired precision on the 
bounds can be attained with some finite values of the basic parameters, performing a 
finite number of iterations. 
The total computational cost per iteration grows linearly with problem size.
The algorithm estimates the marginal density of each separate variable, which
makes it
suitable for parallel implementation.
These features make the proposed method a promising tool, opening the 
possibility of constructing probabilistic optimal certificates for 
large scale unstructured problems. Experimentation in this direction
is in progress. On the other hand, the density estimation algorithm may be
used to develop new heuristics or improve existing stochastic or deterministic
algorithms.

\section*{Acknowledgments}
This work was partially supported by the National Council of Science and Technology of
Mexico under grant CONACYT J45702-A.

\end{document}